\documentclass[sigconf]{acmart}
\AtBeginDocument{%
  \providecommand\BibTeX{{%
    \normalfont B\kern-0.5em{\scshape i\kern-0.25em b}\kern-0.8em\TeX}}}

\acmYear{}
\acmDOI{}
\acmISBN{}

\setcopyright{none}
\acmConference[ICMR '21]{Proceedings of the 2021 International Conference on Multimedia Retrieval}{August 21--24, 2021}{Taipei, Taiwan}
\acmBooktitle{Proceedings of the 2021 International Conference on Multimedia Retrieval (ICMR '21), August 21--24, 2021, Taipei, Taiwan}

\usepackage{graphicx}
\usepackage{amsmath}
\usepackage{amsthm}
\usepackage{booktabs}
\usepackage{algorithm}
\usepackage{algorithmic}
\usepackage{subfigure}
\usepackage{multirow}
\usepackage{tabularx}
\usepackage{hhline}
\newcolumntype{C}{>{\centering\arraybackslash}p{2.8em}}

\def\@fnsymbol#1{\ensuremath{\ifcase#1\or *\or \dagger\or \ddagger\or
   \mathsection\or \mathparagraph\or \|\or **\or \dagger\dagger
   \or \ddagger\ddagger \else\@ctrerr\fi}}
\newcommand{\ssymbol}[1]{^{\@fnsymbol{#1}}}

\usepackage{bm}

\begin{document}

\title{Leveraging EfficientNet and Contrastive Learning for Accurate Global-scale Location Estimation}


\author{Giorgos Kordopatis-Zilos, Panagiotis Galopoulos, Symeon Papadopoulos, Ioannis Kompatsiaris}
\affiliation{%
  \institution{Information Technologies Institute, CERTH, Thessaloniki, Greece}
  \country{}
}
\email{{georgekordopatis,gpan,papadop,ikom}@iti.gr}





\renewcommand{\shortauthors}{Anonymous, et al.}

\begin{abstract}
In this paper, we address the problem of global-scale image geolocation, proposing a mixed classification-retrieval scheme. Unlike other methods that strictly tackle the problem as a classification or retrieval task, we combine the two practices in a unified solution leveraging the advantages of each approach with two different modules. The first leverages the EfficientNet architecture to assign images to a specific geographic cell in a robust way. The second introduces a new residual architecture that is trained with contrastive learning to map input images to an embedding space that minimizes the pairwise geodesic distance of same-location images. For the final location estimation, the two modules are combined with a search-within-cell scheme, where the locations of most similar images from the predicted geographic cell are aggregated based on a spatial clustering scheme. Our approach demonstrates very competitive performance on four public  datasets, achieving new state-of-the-art performance in fine granularity scales, i.e., 15.0\% at 1km range on Im2GPS3k.
\end{abstract}




\maketitle

\section{Introduction}

We define visual location estimation as the process of estimating the geographical coordinates of a scene based solely on the visual cues existing in the given image. One could think multiple variations of this task, depending on whether we restrict our input to a particular kind of scenes, e.g., landmarks \cite{avrithis2010retrieving,Boiarov2019,weyand2020google}, to be from a particular area \cite{wang2020online,kendall2015posenet,arandjelovic2016netvlad}, or on whether we are using different inputs, e.g., a sequence of images per scene \cite{arroyo2015towards,nowicki2016experimental,Nowicki2017}, or  aerial imagery \cite{vo2016localizing,liu2019lending,regmi2019bridging}. In this study, we focus on global-scale location estimation from single images, which is the most challenging problem setting. Without restricting the type and location of the input images, some ambiguity is unavoidable, as not all images contain enough visual cues to allow their precise localization. This ambiguity raises the difficulty of the task significantly and highlights the need to design a general model, resilient to over-fitting, able to extract the informative characteristics from the depicted scenes.

There are two prevalent formulations of the problem of global-scale location estimation and accordingly two  solutions to tackle it: classification and retrieval. The former considers a classification task \cite{Weyand2016,Eric2018,seo2018,Izbicki2019}, partitioning the earth's surface into a grid of cells, and then trains a classifier to assign input images to a grid cell. The latter considers location estimation as a retrieval task \cite{Hays2008,hays2015large,kordopatis2016placing,munozrecod,Vo2017,li2017geo}, searching in a large-scale \emph{background database} of geotagged images to retrieve similar ones based on a given query, and then aggregates them for estimating the location of the query image. Both formulations have limitations. The major drawback of classification approaches stems from the division of the earth into large geographic areas, which results in coarse estimations. A partial solution to that would be to consider a finer granularity grid, but this could potentially hurt the system's performance \cite{Weyand2016,Vo2017,Izbicki2019}. On the other hand, retrieval-based approaches perform worse than classification-based ones \cite{Weyand2016,Eric2018,Izbicki2019}, and they need significantly more computational resources during inference. Also, our investigations indicated that they are prone to noise from the appearance of visually similar concepts within images that are not related to the particular location (e.g., humans, animals, vehicles).

Motivated by the above limitations, we propose a scheme that achieves high geolocation accuracy in all granularity scales. We build on the strong aspects of both classification and retrieval approaches. Their combination has already been employed in global-scale text-based geolocation solutions \cite{van2011finding,kordopatis2017geotagging}; yet, to the best of our knowledge, they have not been successfully employed in the visual domain. We leverage recent advances in the field of image classification, employing a state-of-the-art architecture, and contrastive learning \cite{chen2020simple,he2020momentum,khosla2020supervised}, building a retrieval module that achieves better performance than using features directly extracted from pre-trained CNNs. In particular, we make the following contributions: 
\begin{itemize}
\item
We develop robust classification modules based on the state-of-the-art EfficientNet \cite{tan2019efficientnet} architecture, which has not been employed before in the relevant literature, trained with three different training schemes from the literature. 
\item We build a retrieval module based on a residual architecture trained with contrastive learning. The network learns to capture location-relevant information and enriches the image features representations extracted from the CNN.
\item We also propose the Search within Cell scheme that combines the two modules and estimates the final locations with an aggregation scheme based on spatial clustering.
\item Our approach outperforms several state-of-the-art methods on four benchmark datasets, achieving up to 42\% relative improvement at the 1km range on the Im2GPS3k. We also evaluate our method with various configurations to gain insight into its behaviour.
\end{itemize}

\section{Related Work}
\label{sec:related-work}

There are several works in the literature that tackle the problem of location estimation. These can be roughly classified into two categories according to \cite{brejcha2017state, Eric2018, masone2021survey}: (i) approaches restricted to specific environments or imagery, and (ii) planet-scale approaches without any restrictions. Our approach belongs to the second category. 

The works in the first category focus on the localization on fine granularity scales, such as landmarks \cite{avrithis2010retrieving,Boiarov2019,weyand2020google,yokoo2020two} or at city-scale granularity \cite{wang2020online,kendall2015posenet,arandjelovic2016netvlad,torii2019large, liu2019stochastic}.  In general, the solutions that are employed for such problems are based on retrieval systems that match the query images with ones from a background collection and then apply a post-processing scheme to estimate the final location. However, these methods use restricted data from popular scenes and urban environments and require many instance matches to perform robustly, which is infeasible at a global scale. Another instance that falls within this category are methods that estimate image locations from cross-view imagery, e.g., ground-to-aerial \cite{lin2013cross,vo2016localizing,shi2019spatial,liu2019lending,regmi2019bridging,zhu2021revisiting}. Such methods are restricted on the type of imagery needed as input in order to perform location estimation. 

The works in the second category tackle the location estimation problem under no constraints. Hays et al. \cite{Hays2008} first introduced the problem with the composition of a dataset of about 6 million images collected from Flickr. They proposed an image retrieval method based on handcrafted features. A breakthrough was made by \cite{Weyand2016} when the authors formulated the problem as a classification one and trained a CNN, namely PlaNet, with the cross-entropy loss. The classes were defined using a heuristic process for the adaptive partitioning of the earth into geographic cells. Motivated by the PlaNet \cite{Weyand2016}, a revision of the original Im2GPS paper was made in \cite{Vo2017}, where the authors proposed a retrieval approach for inference, extracting image features from a trained CNN. They also experimented with Deep Metric Learning for fine-tuning the network, yet without achieving significant performance gains. CPlaNet, a modification of the original PlaNet \cite{Weyand2016}, was proposed by \cite{seo2018}. The authors used multiple coarse partitions of the earth and trained a different network for each of them. The final finer-granularity result was calculated by a \emph{combinatorial partitioning} approach, which considered the intersections of the partitions. The authors of \cite{Eric2018} experimented with different architectures, simultaneously using multiple cross-entropy loss functions corresponding to a coarse, middle, and fine-grained partition of the earth for the training of the CNN, and a fusion scheme of the probabilities in the three granularities for the inference of the estimated location. Additionally, they proposed as an initial step to split the images according to the scene they represent, such as indoor, natural or urban, and trained a different model per category. More recently, the authors of \cite{Izbicki2019} proposed the use of the continuous von Mises-Fisher (vMF) distribution to model the geolocation problem as an alternative to the simple classification approach. However, to the best of our knowledge, there has been no global-scale location estimation approach that successfully combines classification and retrieval. Also, the only recently proposed retrieval-based approach \cite{Vo2017} did not manage to achieve better performance compared to using the features extracted from a pre-trained classification network. 

\begin{figure*}[t]
    \centering
    \includegraphics[width=15.5cm]{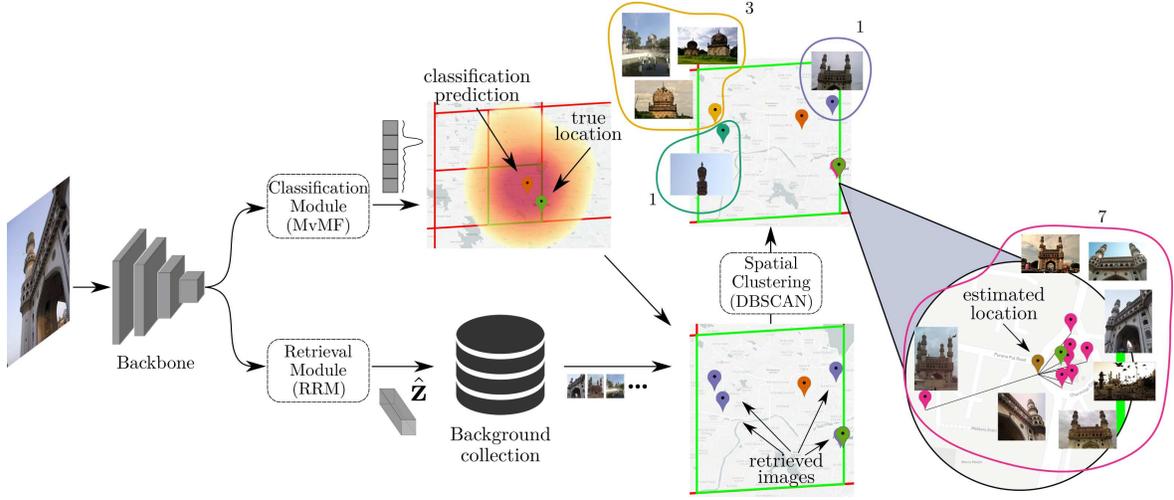}
    \caption{Overview of the proposed Search within Cell (SwC) scheme. Given a query image, a backbone network is first used to extract representative features. Then: (i) a Classification Module predicts a cell on the earth's surface, and (ii) a Retrieval Module extracts an image embedding. These are combined with the Search within Cell scheme, where the most similar images that belong to the predicted cell are retrieved. The final location is estimated based on a Spatial Clustering scheme, where the most similar images are clustered based on their GPS coordinates.}
    \label{fig:methods_overview}
\end{figure*}

\section{Methodology}
\label{sec:methodology}

Figure \ref{fig:methods_overview} illustrates the proposed approach. Our method requires the partitioning of the earth's surface into cells (Section \ref{ssec:partitioning}), which is the base of a classification and a retrieval module. We experiment with three practices to build the classification module (Section \ref{ssec:classification}). For the retrieval module, we propose a residual architecture and train it with contrastive learning to map images to an embedding space, where the same location images have large similarity (Section \ref{ssec:retrieval}). Finally, we combine the two modules with a search within cell scheme and a spatial clustering aggregation approach (Section \ref{ssec:search_within_cell}).

\subsection{Earth Partitioning}
\label{ssec:partitioning}

As the probability distribution of the image locations is not uniformly distributed over the earth, it would make sense to create an adaptive partition of the Earth based on the training data. Similar to \cite{Weyand2016, Eric2018}, the steps of the process we follow are:
(1) create a first coarse partition of the Earth, (2) assign to each cell the included images, (3) choose the cell with the most images and split it, (4) remove the parent cell and assign to each child cell the images located within its borders, (5) repeat steps 3-4 until the desired number of classes is reached. The above process requires a hierarchical representation of the Earth's surface; we used Google's S2 Geometry Library \cite{google2018s2} similar to \cite{Weyand2016, Eric2018}. It is also possible to choose different termination criteria, e.g., terminate when all classes have less images than a fixed threshold, or enforce additional restrictions, e.g., disallow classes to have less than a fixed number of images. For each cell of the resulting partition, we define its center that will be used as a point prediction, estimated as the average location of the image locations in the cell.

\subsection{Classification module}
\label{ssec:classification}

We describe three ways the classification module can be implemented, for which we provide experimental results in Section \ref{sec:results}.

\subsubsection{Discrete probability model trained with cross-entropy}
\label{sssec:simple-classifier}

The most straightforward way to implement the classification module is with a CNN that outputs a discrete probability distribution over the cells defined in Section \ref{ssec:partitioning}, as done by PlaNet \cite{Weyand2016}. The CNN weights can then be trained with the cross-entropy loss, which is commonly used in classification tasks. 

\subsubsection{Hierarchical Classification (HC)}

Another possibility is to follow a hierarchical approach \cite{Eric2018} and simultaneously train three different classifiers, each implemented as described in Section \ref{sssec:simple-classifier}, at different geographical resolutions. That is, we create three different partitions of the Earth, ranging from coarse-grained with a small number of cells to fine-grained with a large number of cells, and attach three different classification heads to the backbone CNN, one per partition. The loss is then calculated as the average of the cross-entropy loss of each head.

Training the backbone with this loss, as done in \cite{Eric2018}, could potentially allow the network to achieve greater generalization power. However, due to the high computational cost, we fix the backbone network weights and train only the different classification heads, which is equivalent to training three independent models at different partitions. For the inference, we can take advantage of the multiple heads by combining the three outputs. Specifically, we 1) calculate the output of each head, 2) find for every cell of the fine partition its parent in the mid and coarse partitions, 3) multiply the probability of the cell with that of its parents, 4) select the cell of the fine partition with the highest probability.

\subsubsection{MvMF trained with log-likelihood loss}
\label{sssec:sph_normal}

A shortcoming of the two previous methods is that they do not take into account that the data points and classes are defined on the surface of the Earth, and as such are related to each other in a common spherical coordinate system. An alternative is to model the task in the continuous probability space using the von Mises-Fisher (vMF) probability distribution \cite{Izbicki2019} that is specifically targeted towards modeling spherical data. The vMF distribution is defined as
\begin{equation}
    \mathrm{vMF}(\bm{x} \mid \bm{\mu}, \kappa) = \frac{\kappa}{4\pi\sinh{\kappa}}e^{\kappa\bm{\mu}^T\bm{x}} \label{eq:vMF}
\end{equation}
where $\lVert \bm{\mu} \rVert = \lVert \bm{x} \rVert = 1$ is the mean and $\kappa > 0$ the concentration, and was used by the authors of \cite{Izbicki2019} to build a probabilistic model of the geolocation task based on a Mixture of vMF distributions (MvMF) that would be trained with the log-likelihood loss: 
first, a partitioning of the Earth is constructed, as described in section \ref{ssec:partitioning}, and each cell of this partitioning corresponds to a single component of the MvMF with mean the center of the contained images. Then, the probability that a given image $I$ is located at $x$ is given by
\begin{equation}
    \mathrm{MvMF}(x \mid I) = \sum_{i=1}^N w_i(I)\mathrm{vMF}(x \mid \bm{\mu}_i, \kappa_i)
\end{equation}
where $\bm{w}(I)$ sums up to one and is calculated from the output of a CNN. The concentration, $\kappa$, of each vMF component is a parameter of the model that is learned during training, but is fixed and independent of the input image $I$ during inference. The network is trained with the negative log-likelihood loss, a common choice for training probabilistic mixture models.

As the final result, ideally, we would choose the location that maximizes the probability density function. However, this would involve solving a computationally intensive non-convex optimization problem; instead, we opt for approximating it with the location of the mean of the mixture component with the highest weight $w$.

\subsection{Retrieval module}
\label{ssec:retrieval}

A robust retrieval system has to map the images in the dataset to an embedding space where the images from the same location are closer to each other than the rest. The goal is to alleviate the high inter-class ambiguity and intra-class diversity introduced by the weak supervision from training with GPS coordinates and noisy data. To this end, we build a network architecture and train it with supervised contrastive learning.

\begin{figure}[t]
    \centering
    \includegraphics[width=4.78cm]{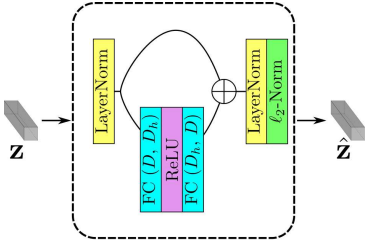}
    \caption{Overview of the Residual Retrieval Module.
    The symbol $\oplus$ indicates element-wise summation.}
    \label{fig:retrieval_network}
\end{figure}

\subsubsection{Network architecture}
\label{ssec:retrieval_network}

To build our retrieval system, we employ a backbone CNN  training based on a classification scheme, and we extract feature representations for the images in our dataset. Given an input image, we feed it to the CNN and apply Global Average Pooling (GAP) on the activations of the final convolutional layer to extract a feature vector $\textbf{z}\in\mathbb{R}^D$, where $D$ is the dimensionality of the output vectors. Inspired by the Feed-Forward Layer from \cite{vaswani2017attention}, we build a residual network for the projection of the image vectors to the embedding feature space. Figure \ref{fig:retrieval_network} displays the network architecture of the retrieval module. The network comprises two fully-connected layers, with $D_h$ hidden size and Rectified Linear Unit (ReLU) \cite{krizhevsky2012imagenet} activation between them. The output is added with the input feature vector with a residual connection. Before and after the residual connection, feature vectors are normalized with LayerNorm \cite{ba2016layer}. Finally, $\ell_2$-normalization is applied on the feature vectors to transform them to the unit sphere. The network output is an enriched embedding vector, $\hat{\textbf{z}}\in\mathbb{R}^D$, with the same dimensionality as the input. With this architecture, our model is able to capture the relevant information from the feature vectors and enrich the image representations without drastically altering them. This is achieved with the use of the residual connection that we empirically found to boost the performance of the proposed system. Our main intuition for the proposed scheme is that the image vectors are already good representations and that the network learns to extract and amplify useful information in the output embeddings.  

\subsubsection{Training process}
\label{ssec:retrieval_training}

For training, we propose a supervised contrastive learning method \cite{khosla2020supervised} based on the image locations. We aim to map the images captured from the same location closer in the embedding space than the rest. Hence, given the Earth's partitioning as described in Section \ref{ssec:partitioning}, images that belong to the same geographic cell are considered as positive pairs; instead, images from different cells are considered as negatives. Let $\hat{\textbf{z}}_q$, $\hat{\textbf{z}}_p$, and $\hat{\textbf{z}}_n^i$, $i=1, 2, ..., N-2$, be the embeddings, derived from our network, of a query image, a positive image (an image from the same cell with the query), and several negative images (images from different cells). We train our network so that the similarity between the query-positive pair is significantly larger than the similarities between the query-negatives. Since all vectors are $\ell_2$-normalized, their dot product measures the similarity between images in the embedding space. To train our network, we employ the infoNCE \cite{oord2018representation} contrastive loss function, as follows:
\begin{equation}
    \mathcal{L}_{nce} = -\log \frac{exp(\hat{\textbf{z}}_q \cdot \hat{\textbf{z}}_p / \tau)}{exp(\hat{\textbf{z}}_q \cdot \hat{\textbf{z}}_p / \tau) + \sum_{i=1}^{N-2} {exp(\hat{\textbf{z}}_q \cdot \hat{\textbf{z}}_n^i / \tau)}}
\end{equation}
where $\tau$ is a temperature hyperparameter \cite{wu2018unsupervised}. The training loss drops as the similarity between the query-positive pair is significantly greater than the similarities of the query and all negatives. Therefore, by minimizing this loss, we force the network to assign high similarity values on the positive pairs, i.e., the images originating from the same cell, and low similarity values on the negative pairs, i.e., the images that come from different cells. To utilize more negative samples during the loss calculation, we employ a cross-batch memory bank \cite{wu2018unsupervised,wang2020cross}. Additionally, to eliminate the bias from the cells that contain many images, in each training epoch, we sample one image pair from each cell. In that way, all cells are equally represented during the training process.

\subsubsection{Background collection}
\label{ssec:retrieval_reference}

In a retrieval system, it is essential to build a background collection where retrieval is performed. In the current problem, the selection of a representative background collection can affect the system performance, as pointed out in \cite{Vo2017}. However, a huge background collection significantly increases the total time needed for retrieval. In this work, we build the background collection with images from the training set, and we populate it with those that our classifier is able to place in the correct geographic cell. In that way, we compose a background collection of ``placeable'' images, i.e., suitable images that serve our retrieval scheme for the visual location estimation task.

\subsection{Search within Cell}
\label{ssec:search_within_cell}

We employ an aggregation scheme to combine the two geolocation modules, classification and retrieval, called Search within Cell (SwC). First, we perform classification to derive the cell with the largest probability. 
Then, we retrieve the top-most similar images of the background collection, with the constraint that the retrieved images fall within the borders of the estimated cell.

Finally, to enhance the robustness of the results, we develop a density-based spatial clustering scheme. Given a query image with its predicted cell, we first retrieve the top $K$ most similar images from the background collection that belongs to the same cell. Our intuition is that the most visually similar images are the most appropriate to infer the location of the query. Then, we apply the DBSCAN \cite{ester1996density} algorithm on the GPS coordinates of the $K$ images to form clusters based on their spatial proximity. DBSCAN is selected because it does not require setting a predefined number of clusters, which would be impractical to select optimally at global scale. We use the geodesic distance as the function to calculate the distance between images. DBSCAN requires setting an $\varepsilon$ threshold, which corresponds to the maximum distance between two samples for the first to be considered in the neighborhood of the second. We empirically set $\varepsilon$ equal to 1km, as we found it to yield marginally better results. Also, DBSCAN may receive as argument the minimum number of samples in a neighborhood for a point to form a cluster. Since we do not want to force the merging of isolated images in the clusters during the final location estimation, we set this parameter to $1$. In the end, the largest cluster (or the first one in rare cases of equal size) is selected. The final location estimation derives from the mean of the locations of the cluster's images.

\section{Evaluation setup}
\label{secevaluation_setup}

\subsection{Datasets}
\label{sec:dataset}
\textbf{MediaEval Placing Task 2016 (MP16)} \cite{choi2016placing} is used for training and evaluation of our approach as provided by the original authors. It consists of approximately 5.8 million geotagged images randomly selected from the YFCC100m \cite{thomee2016yfcc100m} collection, without performing any filtering based on the image metadata. The dataset is split into two parts: (i) the training set, composed of ~4.5 million images, which we use to train our models, and (ii) the test set, including the remaining ~1.3 million images, which is used for evaluation.

\textbf{Im2GPS} \cite{Hays2008} is used for comparison against existing methods as provided by the original authors.  It consists of 237 images from the original Im2GPS dataset, manually selected by the authors of \cite{Hays2008} from a set of 400 images based on their localizability. 

\textbf{Im2GPS3k} \cite{Vo2017} is also used for 
comparison against existing methods as provided by the original authors. It consists of 3,000 images from the original Im2GPS dataset. The images were not manually filtered; hence, it is a more challenging test compared to the previous one. 

\textbf{YFCC4k} \cite{Vo2017} is used for 
comparison against existing methods as provided by the original authors. It consists of 4,000 images from the YFCC100m \cite{thomee2016yfcc100m} dataset without applying any filtering. Since it derives from a general purpose dataset, its image distribution is different from Im2GPS, making it more challenging. 

\subsection{Implementation details}
\label{ssec:implementation}

For the proposed approach, we train an EfficientNet-B4 model \cite{tan2019efficientnet}, initialized with pre-trained weights on ImageNet \cite{deng2009imagenet}. For the training of the backbone CNN, we employ the simple cross-entropy scheme using 32K cells, which we consider as our re-implementation of PlaNet. The model is trained for 25 epochs with the entire training dataset after removing all the images of the users that appeared in the evaluation sets, according to \cite{Vo2017,Eric2018}. We train the network with Stochastic Gradient Descent (SGD) and step decay, with 0.01 initial learning rate, 0.5 decay factor with step 5 epochs, momentum 0.9, batch size 64, and weight decay 10$^{-4}$. During training, we apply data augmentation by randomly cropping image areas that cover at least 70\% of the original images with an aspect ratio in the range from 3/4 to 4/3. The input images are randomly flipped and resized to 300$\times$300 pixels. During inference, we simply resize the images so that the largest side is 300 pixels. For validation, we use the same YFCC100m \cite{thomee2016yfcc100m} subset as in \cite{Eric2018}. After this training session, the weights of the CNN remain fixed and are not updated during the training of the rest of the modules.

For the development of the other two classification schemes, i.e., HC and MVMF, we use similar training processes with the one described above. For HC, we follow the earth partitioning proposed in the original approach \cite{Eric2018}. For the MvMF, we use the same number of cells as in the previous setup. Also, we initialize the weights of the mixture layer with the ones from the classification layer. We train both methods for 5 epochs with the AdamW \cite{loshchilov2018decoupled} optimizer for faster convergence, with $10^{-5}$ initial learning rate, and a cosine annealing learning rate \cite{loshchilov2016sgdr} scheduler. We use the same batch size, weight decay, and augmentation process as above.

\begin{table*}[h]
  \centering
  \setlength\tabcolsep{2.8pt}
    \centering
    \scalebox{0.88}{
    \begin{tabular}{|l|c|cccccccc|cccccccc|}
        \hline
        \multirow{2}{*}{\textbf{Method}} & \multirow{2}{*}{\textbf{Type}} & \multicolumn{8}{c|}{\textbf{Acc@ Im2GPS}} & \multicolumn{8}{c|}{\textbf{Acc@ Im2GPS3k}}\\ 
         &  & \textbf{100m} & \textbf{1km} & \textbf{5km} & \textbf{10km} & \textbf{25km} & \textbf{200km} & \textbf{750km} & \textbf{2500km} & \textbf{100m} & \textbf{1km} & \textbf{5km} & \textbf{10km} & \textbf{25km} & \textbf{200km} & \textbf{750km} & \textbf{2500km}\\ \hline
        \textbf{HC} \cite{Eric2018}      & C & - & 15.2 & - & - & 40.9 & 51.5 & 65.4 & 78.5 & - & 9.7 & - & - & 27.0 & 35.6 & 49.2 & 66.0 \\
        \textbf{ISN} \cite{Eric2018}     & C & - & 16.9 & - & - & 43.0 & 51.9 & 66.7 & 80.2 & - & 10.5 & - & - & 28.0 & 36.6 & 47.7 & 66.0 \\ 
        \textbf{MvMF} \cite{Izbicki2019} & C & - &  8.4 & - & - & 32.6 & 39.4 & 57.2 & 80.2 & - & - & - & - & - & - & - & - \\ 
        \textbf{PlaNet}$\ssymbol{2}$ \cite{Weyand2016} & C & - & 11.0 & 23.6 & 26.6 & 31.2 & 37.6 & 64.6 & 81.9 & - & 8.5 & 18.1 & 21.4 & 24.8 & 34.3 & 48.8 & 64.6 \\
        \textbf{CPlaNet} \cite{seo2018}  & C & - & 16.5 & 29.1 & 33.8 & 37.1 & 46.4 & 62.0 & 78.5 & - & 10.2 & 20.8 & 23.7 & 26.5 & 34.6 & 48.6 & 64.6 \\
        \textbf{PlaNet}$\ssymbol{3}$ \cite{Weyand2016} & C & 4.2 & 17.3 & 33.8 & 38.0 & 41.8 & 53.2 & 67.9 & 82.3 & 2.8 & 11.8 & 22.1 & 25.3 & 28.8 & 37.4 & 51.0 & 67.4 \\
        \textbf{HC}$\ssymbol{3}$ \cite{Eric2018}       & C & 1.7 & 13.5 & 28.7 & 32.9 & 39.7 & 54.0 & \underline{68.8} & \textbf{82.7} & 1.8 & 10.1 & 20.1 & 23.9 & 28.4 & 37.9 & 52.0 & \textbf{68.1} \\
        \textbf{MvMF}$\ssymbol{3}$ \cite{Izbicki2019}  & C & 4.6 & \underline{19.8} & 34.2 & \underline{40.1} & \textbf{44.7} & \textbf{55.7} & 67.5 & 81.9 & 3.0 & 13.1 & \underline{23.5} & \underline{26.7} & \underline{29.8} & \textbf{38.0} & \textbf{52.3} & 67.6 \\ \hline
        \textbf{RevIm2GPS} \cite{Vo2017}   & R & - & 14.4 & - & - & 33.3 & 47.7 & 61.6 & 73.4 & - & 7.2 & - & - & 19.4 & 26.9 & 38.9 & 55.9 \\ 
        \textbf{RRM}                     & R & 5.1 & 19.4 & 35.0 & 37.1 & 40.5 & 51.1 & 60.4 & 78.1 & 3.6 & 12.4 & 20.4 & 23.5 & 26.0 & 34.0 & 46.9 & 63.6 \\ \hline
        \textbf{PlaNet$\ssymbol{3}$ + RRM} & SwC & 5.5 & \underline{19.8} & \underline{36.3} & 38.0 & 41.8 & 52.7 & 67.9 & 82.3 & \textbf{4.3} & \underline{13.9} & 23.4 & 26.1 & 29.3 & 37.4 & 51.0 & 67.4 \\
        \textbf{HC$\ssymbol{3}$ + RRM}     & SwC & 5.5 & 18.6 & 34.6 & 36.7 & 41.8 & \underline{55.3} & \textbf{69.2} & \textbf{82.7} & 3.8 & 13.2 & 22.5 & 25.5 & 29.1 & 37.8 & 52.0 & \textbf{68.1} \\
        \textbf{MvMF$\ssymbol{3}$ + RRM}   & SwC & \textbf{6.3} & \textbf{21.9} & \textbf{38.0} & \textbf{40.5} & \underline{44.3} & \underline{55.3} & 67.5 & 81.9 & \underline{4.1} & \textbf{15.0} & \textbf{24.3} & \textbf{27.0} & \textbf{30.0} & \textbf{38.0} & \textbf{52.3} & 67.6 \\ \hline \multicolumn{2}{c}{} \\
    \end{tabular}
    } 
    \scalebox{0.88}{
    \begin{tabular}{|l|c|cccccccc|cccccccc|}
        \hline
        \multirow{2}{*}{\textbf{Method}} & \multirow{2}{*}{\textbf{Type}} & \multicolumn{8}{c|}{\textbf{Acc@ YFCC4k}} & \multicolumn{8}{c|}{\textbf{Acc@ MP16-test}}\\ 
         &  & \textbf{100m} & \textbf{1km} & \textbf{5km} & \textbf{10km} & \textbf{25km} & \textbf{200km} & \textbf{750km} & \textbf{2500km} & \textbf{100m} & \textbf{1km} & \textbf{5km} & \textbf{10km} & \textbf{25km} & \textbf{200km} & \textbf{750km} & \textbf{2500km}\\ \hline
        \textbf{PlaNet}$\ssymbol{2}$ \cite{Weyand2016} & C &  -  & 5.6 & 10.1 & 12.2 & 14.3 & 22.2 & 36.4 & 55.8 & - & - & - & - & - & - & - & - \\
        \textbf{CPlaNet} \cite{seo2018}                & C &  -  & \textbf{7.9} & \textbf{12.1} & \textbf{13.5} & \textbf{14.8} & \textbf{21.9} & 36.4 & 55.5 & - & - & - & - & - & - & - & - \\
        \textbf{PlaNet}$\ssymbol{3}$ \cite{Weyand2016} & C & 1.8 & 6.1 & 9.7 & 11.3 & 13.0 & 21.0 & 36.4 & 56.2 & 2.0 & 7.3 & 11.7 & 13.5 & 15.3 & 22.4 & 36.9 & 56.1 \\
        \textbf{HC}$\ssymbol{3}$ \cite{Eric2018}       & C & 1.1 & 5.7 & 9.0 & 10.9 & 13.1 & 21.6 & 36.6 & 55.4 & 1.3 & 6.3 & 10.7 & 12.6 & 14.8 & 22.5 & 37.2 & \textbf{56.3} \\
        \textbf{MvMF}$\ssymbol{3}$ \cite{Izbicki2019}  & C & 1.9 & 6.8 & 10.9 & 12.6 & \underline{14.4} & \textbf{21.9} & \textbf{37.5} & \underline{56.4} & 2.0 & 7.8 & \underline{12.5} & \underline{14.3} & \textbf{16.1} & \textbf{23.1} & \textbf{37.4} & \textbf{56.3} \\ \hline
        \textbf{FeatFusion} \cite{munozrecod}& R & - & - & - & - & - & - & - & - & 0.9 & 2.4 & 4.0 & 4.6 & 5.2 & 7.3 & 17.2 & 35.4 \\
        \textbf{VGG-PCA} \cite{kordopatis2016placing}  & R & - & - & - & - & - & - & - & - & 1.8 & 5.6 & 7.5 & 8.2 & 8.8 & 12.1 & 22.4 & 40.8 \\
        \textbf{RevIm2GPS} \cite{Vo2017}   & R & - & 2.3 & - & - & 5.7 & 11.0 & 23.5 & 42.0 & - & - & - & - & - & - & - & - \\ 
        \textbf{RRM}                     & R & 2.3 & 6.0 & 9.0 & 10.2 & 11.3 & 16.8 & 30.5 & 49.4 & \textbf{2.9} & 7.5 & 10.2 & 12.1 & 13.4 & 19.0 & 32.3 & 51.6 \\ \hline
        \textbf{PlaNet$\ssymbol{3}$ + RRM} & SwC & \textbf{2.7} & 7.2 & 10.3 & 11.8 & 13.0 & 20.9 & 36.4 & 56.2 & 2.8 & \underline{8.5} & 12.4 & 13.8 & 15.4 & 22.4 & 36.9 & 56.1 \\
        \textbf{HC$\ssymbol{3}$ + RRM}     & SwC & 2.5 & 7.2 & 10.3 & 11.7 & 13.3 & 21.6 & 36.5 & 55.4 & 2.5 & 8.0 & 11.9 & 13.3 & 15.0 & 22.5 & 37.1 & \textbf{56.3} \\
        \textbf{MvMF$\ssymbol{3}$ + RRM}   & SwC & \textbf{2.7} & \textbf{7.9} & \underline{11.3} & \underline{12.9} & 14.3 & \textbf{21.9} & \underline{37.4} & \textbf{56.5} & \textbf{2.9} & \textbf{8.9} & \textbf{13.1} & \textbf{14.5} & \textbf{16.1} & \textbf{23.1} & \textbf{37.4} & \textbf{56.3} \\ \hline
    \end{tabular}
    }
\caption{Accuracy (\%) on all eight granularity ranges of the proposed and state-of-the-art approaches on four public datasets. The second column indicates the type of the method, C stands for classification, R for retrieval, and SwC for search within cell. The best performances are highlighted in bold, and the second-best are underlined. $\ssymbol{2}$ indicates the results of the re-implemented methods by \cite{seo2018}. $\ssymbol{3}$ indicates the results of our re-implemented methods.}
  \label{tab:sota}
\end{table*}

For the training of the retrieval scheme, we only use image features extracted from the CNN as described in Section \ref{ssec:retrieval_network}. The network is trained for 200 epochs with an AdamW \cite{loshchilov2018decoupled} optimizer, with $10^{-5}$ initial learning rate, and a cosine annealing learning rate scheduler. Also, we use a batch size of 64 image pairs and weight decay 10$^{-4}$. The value of $\tau$ is set to 0.05, the size of the bank is 4096, and the hidden size of the network $D_h$ is 4096. Finally, after filtering the images wrongly placed by the classification network and the images of users that appear in the evaluation sets, the background collection amounts to ~700K images. For the SwC scheme, we use the top 10 most similar images for the location estimation.

The training time on 4 Nvidia RTX 2080Ti was one week for the backbone CNN, approximately 8 hours for each classification scheme, and 2 hours for the retrieval scheme. The inference time of our system is 40ms per image.

\subsection{Evaluation metrics}
\label{ssec:evaluation_metrics}

Following \cite{Hays2008, Vo2017,seo2018}, we evaluate  geolocation performance based on the percentage of images that are placed within a predefined granularity range. An image is considered correctly placed when the geodesic distance of the estimated location to the ground truth is lower than the granularity range. The geodesic distance is calculated as the Great Circle Distance (GCD) between the two locations. We consider two sets of granularity ranges: (i) the baseline granularity ranges, including 1km, 25km, 200km, 750km, and 2500km, corresponding roughly to street, city, region, country, and continent granularity level, and (ii) the fine-grained granularity ranges, including 100m, 1km, 5km, and 10km, that evaluate the methods' performance in fine granularities, which are more representative of the actual performance of a useful location estimation system. 

\section{Experiments}
\label{sec:results}

In this section, we report the results of several runs following the proposed methodology. We compare the different versions of the implemented method against several state-of-the-art approaches (Section \ref{ssec:sota}). Also, we provide an ablation study to evaluate the proposed approach under different configurations (Section \ref{ssec:ablation}).

\subsection{Comparison against the state-of-the-art}
\label{ssec:sota}

Table \ref{tab:sota} illustrates the performance of the proposed and state-of-the-art approaches on the four evaluation datasets. The performance is measured in all eight granularity ranges. The proposed approach is compared with several classification approaches, i.e., Hierarchical Classification (HC) and Individual Scene Networks (ISN) from \cite{Eric2018}, Mixture of von Mises-Fisher (MvMF) \cite{Izbicki2019}, the CPlaNet and Planet \cite{Weyand2016} re-implementation from \cite{seo2018}, the retrieval approaches, i.e., RevIm2GPS \cite{Vo2017}, VGG-PCA \cite{kordopatis2016placing}, and the Feature Fusion approach by \cite{munozrecod}. Moreover, we report the performance of our re-implementations for Planet \cite{Weyand2016}, HC \cite{Eric2018}, and MvMF \cite{Izbicki2019} trained based on our setup described in Section \ref{ssec:implementation}. Finally, the performance of our proposed Residual Retrieval Module (RRM) with nearest neighbour search is demonstrated, and it is combined with our re-implemented classification modules with our SwC scheme.

According to the results, our approach achieves superior performance, especially in fine granularity ranges. Comparing our re-implemented classification run with the original ones, it is clear that the use of the EfficientNet-B4 model significantly boosts performance in almost all granularity ranges, even though we did not train the backbone with HC, which could lead to even better performance according to \cite{Eric2018}. Additionally, our SwC scheme considerably improves geolocation accuracy in fine ranges (i.e., <10km) in all evaluation datasets compared to the individual runs, achieving as high as 2\% absolute performance gain at the 1km range. This highlights that the proposed scheme can operate well with various combinations of classification-retrieval systems. Also, SwC hurts the performance of the classification systems only in very rare occasions and in coarse granularity ranges (i.e., >25km). The HC sees the greatest improvement with the application of the SwC since it uses a coarser grid compared with PlaNet and MvMF, leaving more room for improvement. Finally, our RRM method achieves the best results among the retrieval runs utilizing 700K images as the background collection, which is only a small fraction in comparison to other approaches; RevIm2GPS uses 6 million images, and VGG-PCA and Feature Fusion the entire MP16 training set.

In the Im2GPS, the best performing approach in fine granularity ranges is our SwC scheme implemented with MvMF as the classification module and our RRM. More precisely, it achieves 21.9\% at 1km granularity range, which is a relative improvement of almost 30\% of the previous state-of-the-art achieved by the ISN with 16.9\%. It only leads to a marginal drop at the 25km and 200km ranges. Furthermore, our SwC scheme implemented with HC and RRM achieves the best results in the coarser granularity ranges. It also marginally improves the performance of the classification module at 750km. It is noteworthy that our retrieval module outperforms almost all of the classification methods by a significant margin in fine granularity ranges, i.e., <10km.

In the Im2GPS3k, our SwC scheme with MvMF and RRM outperforms all other approaches by a considerable margin in almost all ranges. It outperforms the previous state-of-the-art ISN method at the 1km range by an absolute difference of 4.5\%. The SwC boosts the performance of the classification module in all fine granularity ranges. Our RRM module demonstrates competitive performance, in particular, at 100m and 1km ranges where it has the best, and second-best performance among the individual runs (i.e., classification and retrieval).

Regarding YFCC4k, the SwC run with MvMF+RRM leads to the best results in the finer and coarser ranges, i.e., $\leq$1km and $\geq$200km, and the second-best in the ranges from 5km to 25km behind CPlaNet. However, we empirically found that using images of the users in the evaluation set for training of the classification modules considerably improves performance, i.e., more than an absolute 2\% in any granularity. Nevertheless, it is not clear in \cite{seo2018} whether such images were used during training of the CPlaNet and PlaNet re-implementation. Also, it is worth noting that the performance of all runs is considerably lower, indicating that YFCC4k is much more challenging than Im2GPS. This is expected since it consists of random images from the YFCC100m without any filtering. Such images may not be appropriate for the evaluation of the geolocation problem. Yet, this dataset simulates an unconstrained scenario where any arbitrary image has to be geolocated. 

Finally, the performance on the MP16-test is similar to the YFCC4k since both datasets derive from the same distribution, i.e., they are random samples from the YFCC100m. The SwC run with MvMF+RRM outperforms all others in all granularity ranges. The results of the two retrieval runs were provided by the organizers of the MediaEval Placing Task 2016. Our RRM achieves significantly better results than the previous retrieval approaches, highlighting the progress in the field over the last years.

\begin{figure*}[t]
    \centering
    \subfigure[correct classification - correct retrieval]{\includegraphics[width=8.4cm]{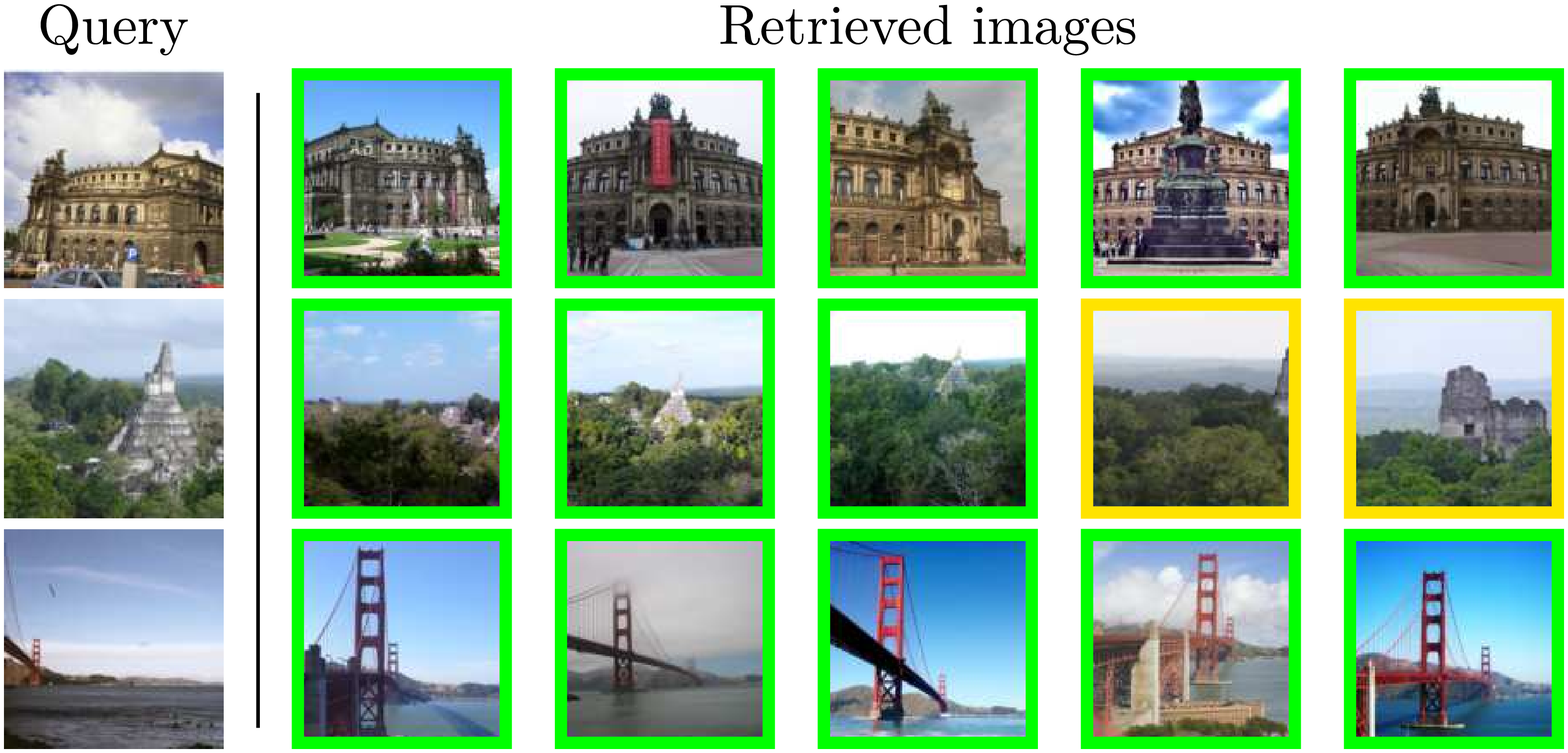}\label{fig:ret_c_clas_c}} \quad
    \subfigure[wrong classification - correct retrieval]{\includegraphics[width=8.4cm]{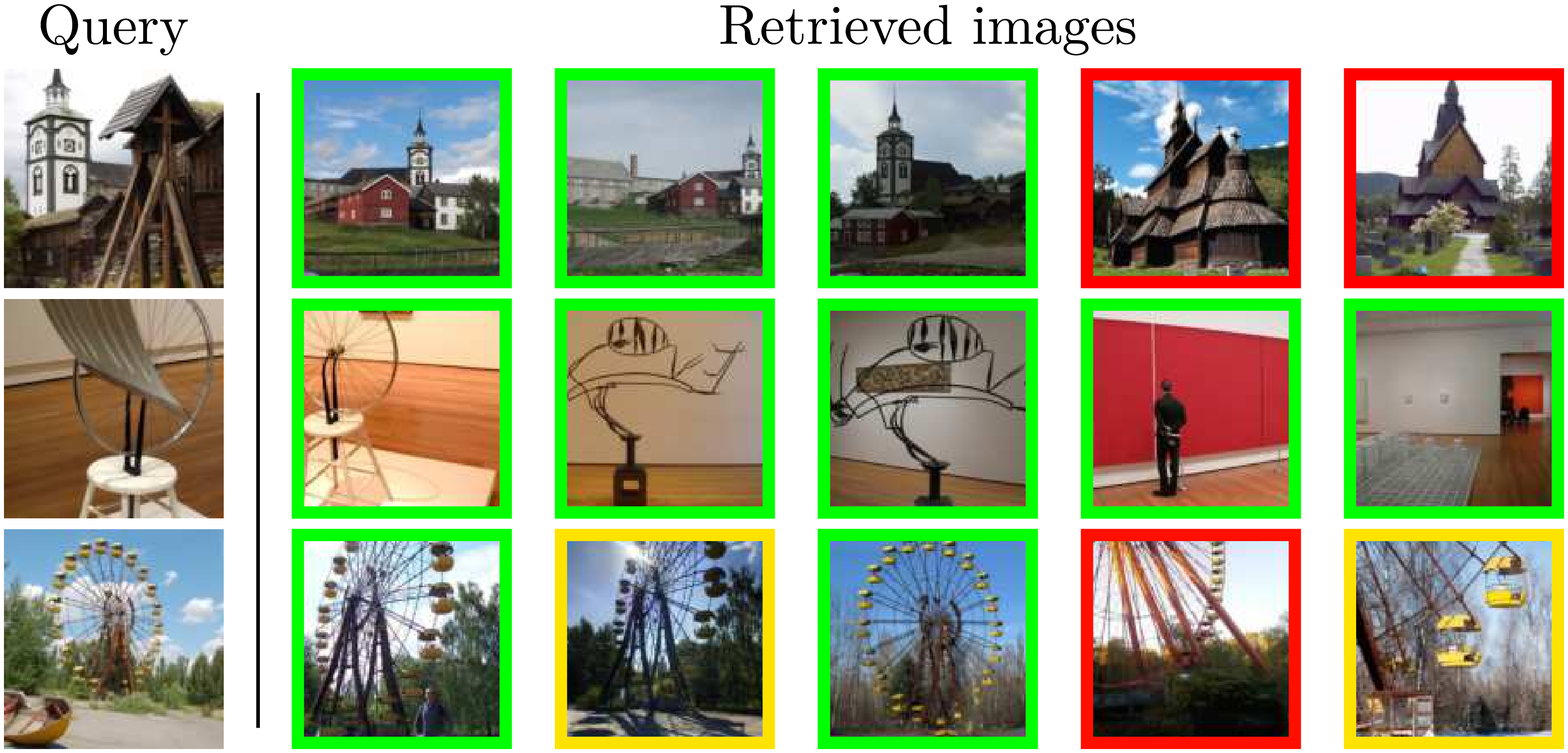}\label{fig:ret_c_clas_w}}
    \subfigure[correct classification - wrong retrieval]{\includegraphics[width=8.4cm]{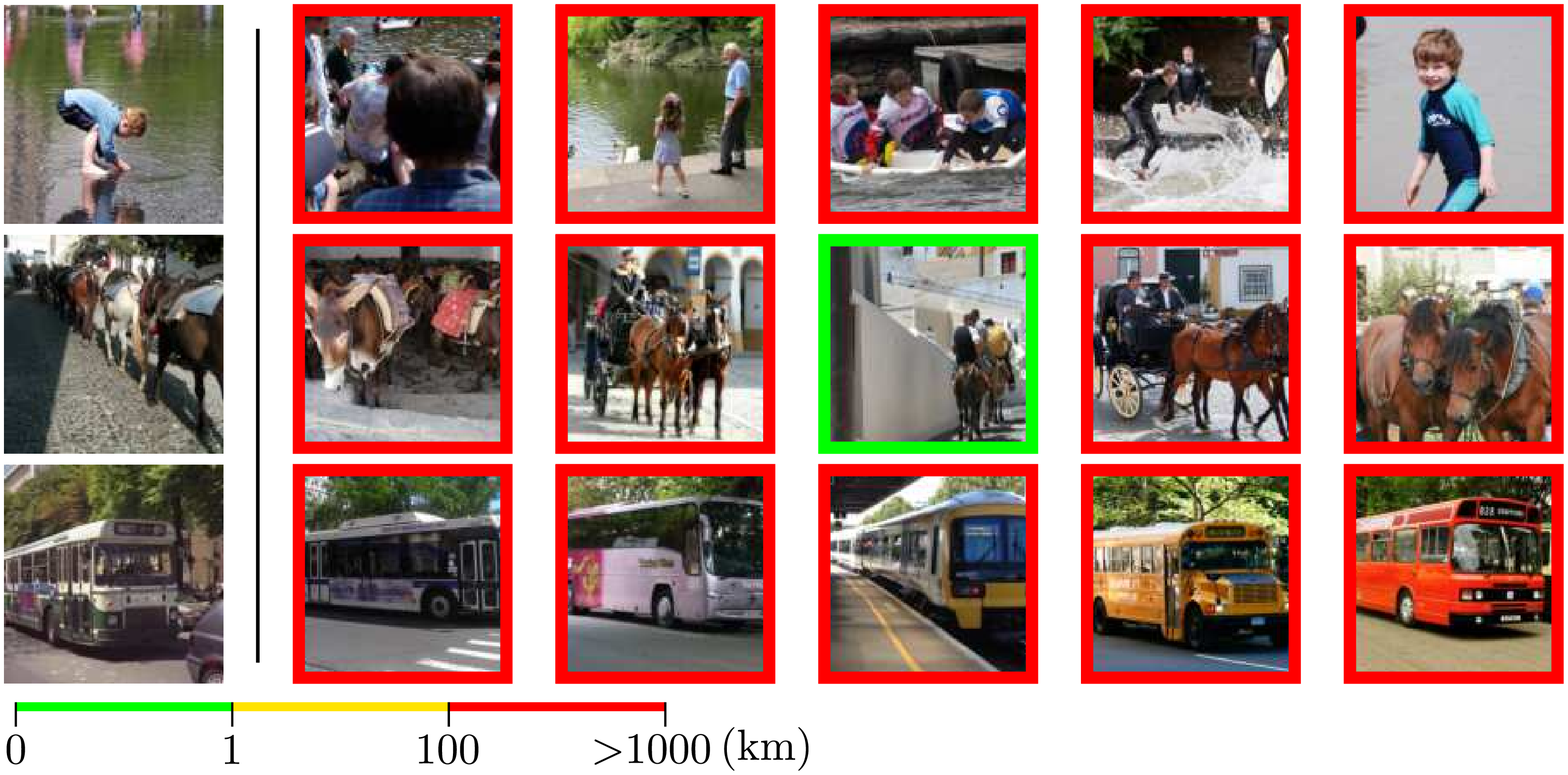}\label{fig:ret_w_clas_c}}  \quad
    \subfigure[wrong classification - wrong retrieval]{\includegraphics[width=8.4cm]{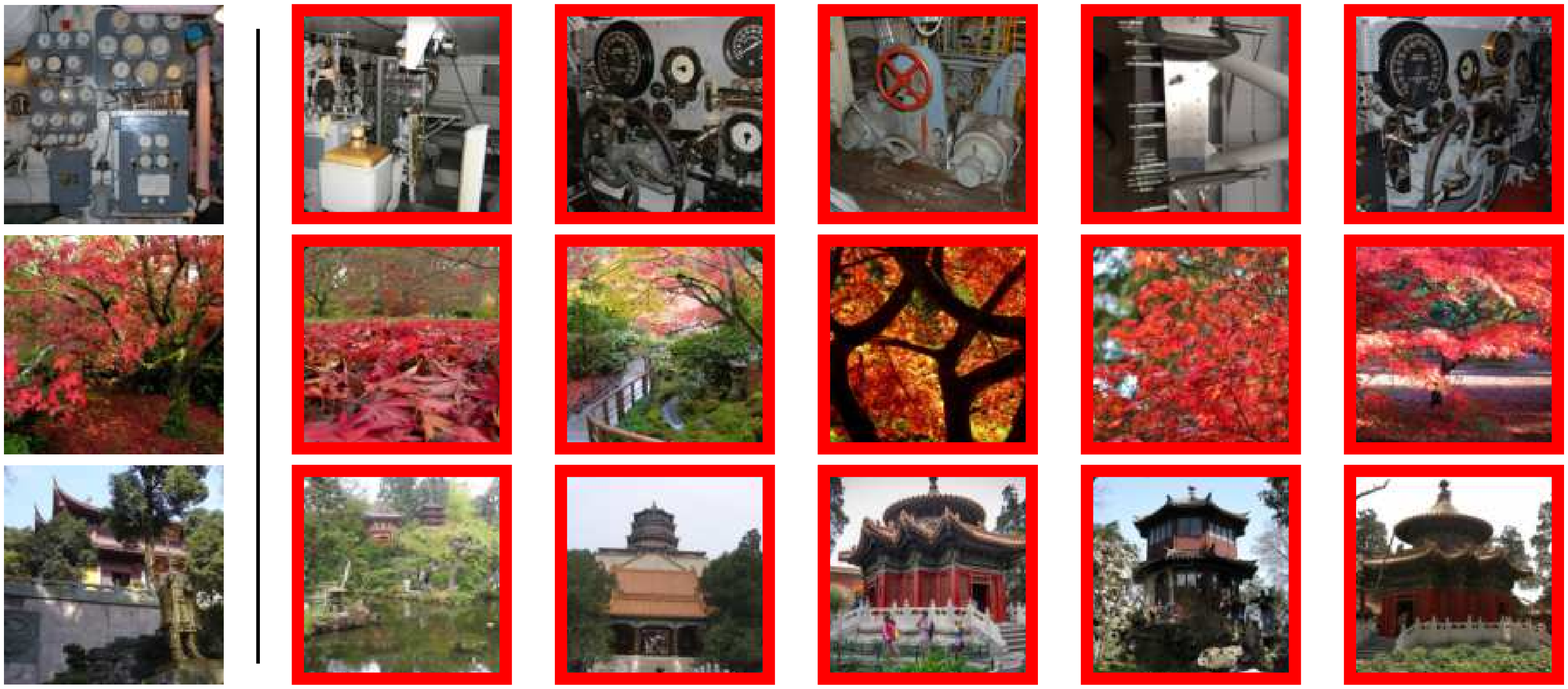}\label{fig:ret_w_clas_w}}
    \caption{Top-5 images from the background collection retrieved by our RRM given the image in the left as query. Retrieved images are coloured based on their distance to the ground truth location of the query: green indicates less than 1km, yellow within 1km and 100km, and red more than 100km. The images are grouped based on the predictions of the RRM and the MvMF classification modules: (a) both modules found the correct location, (b) correct predictions by the RMM and wrong by the MvMF, (c) wrong predictions by the RMM and correct by the MvMF, (d) both modules wrong. (best viewed in color)}
    \label{fig:retrieval_results}
\end{figure*}

\subsection{Ablation study}
\label{ssec:ablation}

This section provides an ablation study on the Im2GPS3k dataset for our proposed RRM module and SwC scheme. We benchmark their geolocation performance to better understand their behavior under different configuration settings.

First, we compare the performance of the proposed retrieval module trained with different loss functions and against a baseline run. For the baseline, the features extracted from the CNN are directly used for retrieval without the application of the retrieval module or any further training. Also, for the training of the RRM, we compare the infoNCE loss against the triplet loss employed in \cite{Vo2017}. We use a margin of 0.01, and semi-hard mining \cite{schroff2015facenet}, which yielded the best performance. Table \ref{tab:retrieval} depicts the results in the baseline granularity ranges. The proposed scheme with infoNCE achieves the best performance in all ranges. It outperforms the baseline by a significant margin. The run with the triplet loss does not lead to competitive results even in comparison to the baseline, in accordance with the observations in \cite{Vo2017}.

\begin{table}[t]
    \centering
    \scalebox{0.89}{
    \begin{tabular}{|l|CCCCc|}
        \hline
        \textbf{loss} & \textbf{1km} & \textbf{25km} & \textbf{200km} & \textbf{750km} & \textbf{2500km} \\ \hline
        \textbf{baseline}    & 11.9 & 24.7 & 31.6 & 43.4 & 59.4 \\
        \textbf{triplet}     & 11.6 & 24.4 & 32.0 & 45.0 & 62.2 \\
        \textbf{infoNCE} & \textbf{12.4} & \textbf{26.0} & \textbf{34.0} & \textbf{46.9} & \textbf{63.6} \\ \hline
    \end{tabular}
    }
    \caption{Accuracy (\%) on the baseline granularity ranges of the baseline and the proposed retrieval module trained with infoNCE and triplet loss on Im2GPS3k.}
  \label{tab:retrieval}
\end{table}

\begin{table}[t]
    \centering
    \scalebox{0.89}{
    \begin{tabular}{|l|CCCCc|}
        \hline
        \textbf{Network} & \textbf{1km} & \textbf{25km} & \textbf{200km} & \textbf{750km} & \textbf{2500km}\\ \hline
        \textbf{w/o residual}    & 11.2 & 24.8 & 32.6 & 45.5 & 62.0 \\
        \textbf{w/ residual}     & \textbf{12.4} & \textbf{26.0} & \textbf{34.0} & \textbf{46.9} & \textbf{63.6} \\ \hline
    \end{tabular}
    }
    \caption{Accuracy (\%) on the baseline granularity ranges of the proposed retrieval module with and without the residual connection on Im2GPS3k.}
  \label{tab:residual}
\end{table}

Additionally, in Table \ref{tab:residual} we evaluate the impact of the residual connection on the geolocation performance of the RRM module. It is evident that the application of the residual connection considerably improves performance, highlighting its importance to the proposed system. Comparing these runs with the baseline one from Table \ref{tab:retrieval}, it appears that the residual connection leads to a clear accuracy increase across all ranges. 

Table \ref{tab:reference} depicts the performance of the RRM using the proposed background collection (700K images) and the entire training set (4M images). With the proposed background collection, we achieve considerably better accuracy at the 1km range with almost 2\% difference; whereas, for most of the other ranges, the use of the entire training set provides marginally better performance. Considering that the proposed collection is only a fraction (17.5\%) of the training set, which translates to much faster retrieval and lower memory requirements, we find that it strikes an excellent trade-off between accuracy and speed.

\begin{table}[t]
    \centering
    \scalebox{0.89}{
    \begin{tabular}{|l|CCCCc|}
        \hline
        \textbf{background col.} & \textbf{1km} & \textbf{25km} & \textbf{200km} & \textbf{750km} & \textbf{2500km}\\ \hline
        \textbf{proposed}    & \textbf{12.4} & 26.0 & 34.0 & 46.9 & \textbf{63.6} \\
        \textbf{all train set}      & 10.5 & \textbf{26.8} & \textbf{34.9} & \textbf{47.2} & 63.5 \\ \hline
    \end{tabular}
    }
    \caption{Accuracy (\%) on the baseline granularity ranges of the proposed retrieval module with different background collection on Im2GPS3k.}
  \label{tab:reference}
\end{table}

\begin{table}[t]
    \centering
    \scalebox{0.89}{
    \begin{tabular}{|c|cccc|}
        \hline
        $K$ & \textbf{100m} & \textbf{1km} & \textbf{5km} & \textbf{10km} \\ \hline
        1     & 3.6 & 14.4 & 24.1 & 26.8 \\
        5     & 4.0 & 14.7 & \textbf{24.3} & \textbf{27.1} \\
        10    & \textbf{4.1} & \textbf{15.0} & \textbf{24.3} & 27.0 \\
        15    & 3.9 & 14.9 & 24.1 & 27.0 \\
        20    & 3.8 & 14.7 & 24.1 & 26.8 \\ \hline
    \end{tabular}
    }
    \caption{Accuracy (\%) on the fine-grained granularity ranges of the proposed SwC scheme with MvMF and RRM modules for different values of $K$ on Im2GPS3k.}
  \label{tab:top_k}
\end{table}

We also investigate the impact of the selection of the temperature $\tau$ hyperparameter. The best performance is achieved when $\tau$ equals 0.05, which drops for greater or lower values (e.g., for 0.1 and 0.01, Acc@1km is 11.8 and 12.1, respectively). We also tested various sizes of the cross-batch memory bank, and  conclude that the larger the size, the better the performance. Due to GPU memory limitations, we experimented with memory size up to 4096 vectors.

Moreover, we assess the impact of the selection of $K$ for the SwC scheme. Table \ref{tab:top_k} presents the accuracy of the approach for various $K$ values in the fine-grained granularity range. Our system achieves the best results for $K=10$ in finer ranges, i.e., 100m and 1km, and for $K=5$ in coarser ranges. For values greater than 10, the performance starts dropping.

\begin{table}[t]
    \centering
    \scalebox{0.9}{
    \begin{tabular}{|l|cccc|}
        \hline
        \textbf{Aggregation} & \textbf{100m} & \textbf{1km} & \textbf{5km} & \textbf{10km} \\ \hline
        \textbf{Average}            & 3.9 & 14.1 & 24.0 & 26.8 \\
        \textbf{KDE} \cite{Vo2017}  & 3.8 & 14.5 & 24.1 & 26.8 \\
        \textbf{Spatial clustering} & \textbf{4.2} & \textbf{15.0} & \textbf{24.3} & \textbf{27.0} \\ \hline
    \end{tabular}
    }
    \caption{Accuracy (\%) on the fine-grained granularity ranges of the proposed SwC scheme with MvMF and RRM modules with different aggregations on Im2GPS3k.}
  \label{tab:aggregation}
\end{table}

Finally, we benchmark three aggregation schemes for the final location estimation. We compare the proposed spatial clustering with a simple averaging of the image coordinates and with Kernel Density Estimation (KDE), as proposed in \cite{Vo2017}. For all schemes, the top-10 similar images are used for the location estimation. Table \ref{tab:aggregation} depicts the results in the fine-grained granularity range. It is evident that the proposed approach achieves the best performance by a considerable margin in all granularity ranges.

\subsection{Qualitative evaluation}
In this section, we provide some visual examples of the retrieved images based on our RRM module, given some queries. The top-5 images are illustrated in colour based on their geodesic distance from the queries' ground truth location. Also, the images are grouped according to the performance of the RRM and MvMF modules. Figure \ref{fig:ret_c_clas_c} displays the images that were placed within 1km from their true location by both methods. It is evident that many visual cues are present, mapping the images to their precise locations. Figure \ref{fig:ret_c_clas_w} presents the queries that were correctly placed by the RRM but missed by the MvMF. There are visual cues in the queries mapping the images to their locations; thus, the retrieval module can find several related images from the same location, highlighting that there is room for improvement in our proposed SwC scheme. Figure \ref{fig:ret_w_clas_c} shows some example queries that were wrongly placed by the retrieval module but correctly placed by the classification module. It is noteworthy that the retrieval module is distracted by the same concepts that appear in both query and reference images, i.e., the child in the first example, the donkeys in the second, and the bus in the third. Such cases are correctly addressed with our SwC scheme, as it confines the RRM to search for similar images within the borders of the cell predicted by the MvMF. Finally, Figure \ref{fig:ret_w_clas_w} illustrates examples of queries that were wrongly placed by both modules. These cases either lack visual cues to map them to their location, i.e., in the first two examples, or cues are too ambiguous mapping the images to multiple locations, i.e., in the third example where many buildings with similar architectural style exist in different locations.

\section{Conclusions}

In this paper, we proposed a method for planet-scale location estimation that combines a classification and a retrieval module to estimate the location of a query image. We built three state-of-the-art classification schemes using EfficientNet \cite{tan2019efficientnet} as backbone and proposed a retrieval module based on a residual architecture trained with contrastive learning. Our method exhibits very competitive performance on four datasets, significantly improving the state-of-the-art in many granularity ranges. In the future, we plan to investigate leveraging text annotations of images during training in order to build more robust classification and retrieval models.

\section*{Acknowledgements} This work has been supported by the We-Verify project, partially funded by the European Commission under contract number 825297.

\balance
\bibliographystyle{ACM-Reference-Format}
\bibliography{refs}

\end{document}